\documentclass{article}

    \PassOptionsToPackage{numbers, compress}{natbib}
    \usepackage[preprint]{neurips_2020}

\usepackage[utf8]{inputenc} % allow utf-8 input
\usepackage[T1]{fontenc}    % use 8-bit T1 fonts
\usepackage{hyperref}       % hyperlinks
\hypersetup{colorlinks=true}
\usepackage{url}            % simple URL typesetting
\usepackage{booktabs}       % professional-quality tables
\usepackage{amsfonts}       % blackboard math symbols
\usepackage{nicefrac}       % compact symbols for 1/2, etc.
\usepackage{microtype}      % microtypography

\usepackage{graphicx}
\usepackage{amsbsy}
\usepackage{color}

\newcommand{\ds}{\displaystyle}
\newcommand{\de}{\textnormal{d}}
\newcommand{\avg}[1]{\left\langle#1\right\rangle}
\newcommand{\abs}[1]{\left|#1\right|}

\newcommand{\set}[1]{\left\{#1\right\}}
\newcommand{\gauss}{\mathcal{N}}

\newcommand{\relu}{\mathcal{R}}
\newcommand{\loss}{\mathcal{L}}
\newcommand{\xvec}{\vec x}
\newcommand{\xhat}{{\bf \hat x}}
\newcommand{\xb}{{\bf x}}
\newcommand{\xpar}{{x^\parallel}}

\newcommand{\wb}{{\bf w}}
\newcommand{\wx}{w_x}
\newcommand{\wplh}{\hat {\bf w}^+}
\newcommand{\wpl}{{w^+}}
\newcommand{\wref}{\hat {\bf w}^*}
\newcommand{\wpar}{w^\parallel}

\newcommand{\wperr}{{w_\perp}}

\newcommand{\erf}{{\rm erf}}
\newcommand{\Reals}{\mathbb{R}}
\newcommand{\Ind}{\mathbb{I}}
\newcommand{\thb}{{\boldsymbol{\theta}}}

\title{An analytic theory of shallow networks dynamics for hinge loss classification}

\author{
  Franco Pellegrini\\
  Laboratoire de Physique de l'\'{E}cole normale sup\'{e}rieure, ENS,\\
  Universit\'{e} PSL, CNRS, Sorbonne Universit\'{e}, Universit\'{e} de Paris\\
  F-75005 Paris, France\\
  \And
  Giulio Biroli\\
  Laboratoire de Physique de l'\'{E}cole normale sup\'{e}rieure, ENS,\\
  Universit\'{e} PSL, CNRS, Sorbonne Universit\'{e}, Universit\'{e} de Paris\\
  F-75005 Paris, France\\
}

\begin{document}
\maketitle
\begin{abstract}
Neural networks have been shown to perform incredibly well in classification tasks over structured high-dimensional datasets. However, the learning dynamics of such networks is still poorly understood. In this paper we study in detail the training dynamics of a simple type of neural network: a single hidden layer trained to perform a classification task.
We show that in a suitable mean-field limit this case maps to a single-node learning problem with a time-dependent dataset determined self-consistently from  the average nodes population.
We specialize our theory to the prototypical case of a linearly separable dataset and 
a linear hinge loss, for which the dynamics can be explicitly solved. This allow us to address in a simple setting several phenomena appearing in modern networks such as slowing down of training dynamics, crossover between rich and lazy learning, and overfitting. 
Finally, we asses the limitations of mean-field theory by studying the case of large but finite number of nodes and of training samples.
\end{abstract}

%%%%%%%%%%%%%%%%%%%%%%%%%%%%%%%%%%%%%%%%%%%%%%%%%%%%%%%%%
\section{Introduction}
Despite their proven ability to tackle a large class of complex problems~\cite{LeCun15}, neural networks are still poorly understood from a theoretical point of view.
While general theorems prove them to be universal approximators~\cite{Barron93}, their ability to obtain generalizing solutions given a finite set of examples remains largely unexplained.
This behavior has been observed in multiple settings. 
The huge number of parameters and the optimization algorithms  employed to optimize them (gradient descent and its variations) are thought to play key roles in it~\cite{Poggio17,Suggala18,Gidel19}.

In consequence, a large research effort has been devoted in recent years to understanding the training dynamics of neural networks with a very large number of nodes \cite{dauphin2014identifying,sagun2016singularity,baity2019comparing}.
Much theoretical insight has been gained in the training dynamics of linear~\cite{Saxe13,Lampinen18} and nonlinear networks for regression problems, often with quadratic loss and in a teacher-student setting~\cite{Saad95,Advani17,Goldt19,Yoshida19}, highlighting the evolution of correlations between data and network outputs. More generally, the input-output correlation and its effect on the landscape has been used to show the effectiveness of gradient descent~\cite{Du18,Arora19}.
Other approaches have focused on infinitely wide networks to perform a mean-field analysis of the weights dynamics~\cite{Mei18,Mei19,Kadmon16,Rotskoff18,araujo2019mean,nguyen2019mean}, or study its neural tangent kernel (NTK, or ``lazy'') limit~\cite{Chizat19,Jacot18,Lee19,geiger2020scaling}.

In this work, we investigate the learning dynamics for binary classification problems, by considering one of the most common cost functions employed in this setting: the linear hinge loss.
The idea behind the hinge loss is that examples should contribute to the cost function if misclassified, but also if classified with a certainty lower than a given threshold. In our case this cost is linear in the distance from the threshold, and zero for examples classified above threshold, that we shall call {\it satisfied} henceforth.
This specific choice leads to an interesting consequence: the instantaneous gradient for each node due to {\it unsatisfied} examples depends on the activation of the other nodes only through their population, while that due to {\it satisfied} examples is just zero.
Describing the learning dynamics in the mean-field limit amounts to computing the effective example distribution for a given distribution of parameters: each node then evolves ``independently'' with a time-dependent dataset determined self-consistently from  the average nodes population.

{\bf Contribution.} We provide an analytical theory for the dynamics of a single hidden layer neural network trained for binary classification with linear hinge loss. 
In Sec.~\ref{sec:MF} we obtain the mean-field theory equations for the training dynamics. Those equations are a generalizations of the ones obtained for mean-square loss in~\cite{Mei18,Mei19,Kadmon16,Rotskoff18,araujo2019mean,nguyen2019mean}. In Sec.~\ref{sec:linsep} we focus on linearly separable data with spherical symmetry and present an explicit analytical solution of the dynamics of the nodes parameters. In this setting we provide a detailed study of the cross-over between the lazy \cite{Chizat19} and rich \cite{woodworth2020kernel} learning regimes (Sec.~\ref{ssec:lazy}).
Finally, we asses the limitations of mean-field theory by studying the case of large but finite number of nodes and finite number of training samples (Sec.~\ref{ssec:finitesize}). The most important new effect is overfitting, which we are able to describe by analyzing corrections to mean-field theory. 
In Sec.~\ref{ssec:mislabel} we show that introducing a small fraction of mislabeled examples induces a slowing down of the dynamics and hastens the onset of the overfitting phase.
Finally in Sec.~\ref{sec:disc} we present numerical experiments on a realistic case, and show that the associated nodes dynamics in the first stage of training is in good agreement with our results. \\
The merit of the model we focused on is that, thanks to its simplicity, several effects happening in real networks can be studied analytically.
Our analytical theory is derived using reasoning common in theoretical physics, which we expect can be made rigorous following the lines of~\cite{Mei18,Mei19,Kadmon16,Rotskoff18,araujo2019mean,nguyen2019mean}. All our results are tested throughout the paper by numerical simulations which confirm their validity. 

{\bf Related works.} Mean-field analysis of the training dynamics of very wide neural networks have mainly focused on regression problems with mean-square losses~\cite{Mei18,Mei19,Kadmon16,Rotskoff18,araujo2019mean,nguyen2019mean,Chizat19}, whereas fewer works~\cite{Combes19,Nacson19} have tackled the dynamics for classification tasks.\footnote{In the NTK (or ``lazy'') limit~\cite{Chizat19,Jacot18,Lee19} general losses have been considered.}
The model of data we focus on bears strong similarities to the one proposed in~\citet{Combes19}, but with fewer assumptions on the dataset and initialization. With respect to~\cite{Combes19}, we show the relation with mean-field treatments~\cite{Mei18,Mei19,Kadmon16,Rotskoff18,araujo2019mean,nguyen2019mean} and provide a full analysis of the dynamics, in particular the cross-over between rich and lazy learning. Moreover, we discuss the limitations of mean-field theory, the source of overfitting and the change in the dynamics due to mislabeling. 

%%%%%%%%%%%%%%%%%%%%%%%%%%%%%%%%%%%%%%%%%%%%%%%%%%%%%%%%%
\section{Mean-Field equation for the density of parameters}\label{sec:MF}
We consider a binary classification task for $N$ points in $d$ dimensions $\set{{\xb}_n} \subset\Reals^d$ with corresponding labels $y_n=\pm1$.
We focus on a hidden layer neural network consisting of $M$ nodes with activation $\sigma$. The output of the network is therefore
\begin{equation}
    f(\xvec;\thb)=\frac{1}{{M}}\sum_{i=1}^{M}
    a_i\sigma\left(\frac{\wb_i\cdot\xb}{\sqrt{d}}\right),
\end{equation}
where $\thb_i=\set{a_i,{\wb}_i}$ represents all the trainable parameters of the model: $\set{\wb_i}$, the $d$-dimensional weight vectors between input and each hidden node, and $\set{a_i}$, the contributions of each node to the output.
All components are initialized before training from a Gaussian distribution with zero mean and unit standard deviation.
%: $\thb_0=\set{a_0,\wb_0}\sim\gauss(0,1)$. 
The $1/M$ in front of the sum leads to the so-called mean-field normalization \cite{Mei18}. In the large-$M$ limit, 
this allows to do what is called a hydrodynamic treatment in physics, a procedure that have been put on a rigorous basis in this context in~\cite{Mei18,Mei19,Kadmon16,Rotskoff18,araujo2019mean,nguyen2019mean,Chizat19} (here the $\thb_i$s play the role of particle positions). In this limit one can rewrite the output function in terms of the averaged nodes population (or density) $\rho(\thb)$:
\begin{equation}\label{eq:f}
    {f(\xb;\thb)=\int d\thb \rho(\thb) }
    a\sigma\left(\frac{{\wb\cdot\xb}}{\sqrt{d}}\right).
\end{equation}

To optimize the parameters we minimize the loss function 
\begin{equation}\label{eq:Loss}
    {\mathcal L}=\frac 1 N \sum_{n=1}^N \ell(y_n,f(\xb_n;\thb))
\end{equation}
by gradient flow $\dot{\thb}=-\beta^*\partial\loss/\partial\thb$ ($\ell(x,y)$ will be specified later). The dynamical equations for the parameters $\set{a_i,\wb_i}$
read: 
\begin{equation}\left\{\begin{array}{rcl}
    \dot a_i&=&\ds -\frac{\beta}{N}\sum_{n=1}^N
    \frac{\partial \ell(y_n,f({\xb;\thb}))}{\partial f}\sigma\left(\frac{{\wb_i\cdot\xb}}{\sqrt{d}}\right)
    \\
    \dot{{\wb}_i}&=&\ds- \frac{\beta}{N}\sum_{n=1}^N\frac{\partial \ell(y_n,f({\xb;\thb}))}{\partial f}a_i
    \sigma'\left(\frac{\wb_i\cdot\xb}{\sqrt{d}}\right)\frac{\xb}{\sqrt{d}},
\end{array}\right.\end{equation}
where we have defined the effective learning rate $\beta=\beta^*/M$.
These equations show that the coupling between the different nodes has a mean-field form: it is through the function $f$, i.e.\ only through the density $\rho(\thb,t)$. Following standard techniques one can obtain in the large $M$ limit a closed hydrodynamic-like equation on $\rho(\thb,t)$  (see Appendix~\ref{ssec:SMhydro} for details): 
\begin{equation}\label{hydro}
\partial_t\rho(\thb,t)=\beta\nabla_{\thb} \left(\rho(\thb,t)\nabla_{\thb}\frac{\delta {\mathcal L[\rho(\thb,t)]}}{\delta \rho(\thb,t)}\right)\,\,,\,\, \rho(\thb,0)=\gauss(0,\mathbb{I})
\end{equation}
where we have made explicit that the $\mathcal L$ is a functional of the density $\rho$ since it depends on $ f(\xb;\thb)$, see eqs.~(\ref{eq:f}, \ref{eq:Loss}). 

To be more concrete, in the following we consider the case of linear hinge loss, $\ell(y,f)=\relu(h-yf)$ ($h$ being the size of the hinge, often taken as 1), and rectified linear unit (ReLU) activation function: $\sigma(x)=\relu(x)={\rm max}(0,x)$. With this choice 
\begin{equation}
    \frac{\delta {\mathcal L[\rho(\thb,t)]}}{\delta \rho(\thb,t)}=
    -a\avg{u(\xb,y;t)\theta\left({\wb\cdot\xb}\right)y\frac{\wb\cdot\xb}{\sqrt{d}}}_{\xb,y} \,.\label{gradl}
\end{equation}
The notation $u(\xb,y;t)\equiv\Ind_{h-yf(\xb;\thb(t))>0}$ denotes the indicator function of the {\it unsatisfied} examples, i.e.\ those $({\xb},y)$ for which the loss is positive, and $\avg{\cdot}_{\xb,y}$ denotes the average over examples and classes ($y=\pm1$ for binary classification). 
The dynamical equations on the node parameters simplify too: 
\begin{equation}\left\{\begin{array}{rcl}
    \dot a_i(t)&=&\ds \frac{\beta}{\sqrt{d}}{\wb}_i\cdot
    \avg{u(\xb,y;t)\theta\left({{\wb}_i\cdot\xb}\right)y\,\xb}_{\xb,y}
    \\[10pt]
    \dot{{\wb}_i}(t)&=&\ds\frac{\beta}{\sqrt{d}}
    \,a_i\,\avg{u(\xb,y;t)\theta\left({{\wb}_i\cdot\xb}\right)y\,\xb}_{\xb,y}.
\end{array}\right.
\label{aweq}\end{equation}
Remarkably, the equation on the ${\wb}_i$ is very similar to the one induced by the Hebb rule in biological neural networks. 

%%%%%%%%%%%%%%%%%%%%%%%%%%%%%%%%%%%%%%%%%%%%%%%%%%%%%%%%%
\section{Analysis of a linearly separable case}\label{sec:linsep}
We now focus on a linearly separable model, where the dynamics can be solved explicitly. 
We consider a reference unit vector $\wref$ in input space and examples distributed according to a spherical probability distribution $P(\xb)$. We label each example based on the sign of its scalar product with $\wref$, leading to a distribution for $y=\pm1$: $P(\xb,y)=P(\xb)\theta(y(\wref\cdot\xb))$.

In order to be able to explore different training regimes, we adopt a rescaled loss function, similar to the one proposed in \citet{Chizat19}:
\begin{equation}\label{eq:loss1}
    \loss^\alpha(\thb)=\frac{1}{\alpha^2 N}\sum_{n=1}^N
    \relu\Big[h-\alpha y_n\Big(f(\xb_n;\thb)
    -f(\xb_n;\thb_0)\Big)\Big],
\end{equation}
where $\alpha$ is the rescaling parameter and $\thb_0$ are the parameters at the beginning of training. Subtracting the initial output of the network ensures that no bias is introduced by the specific finite choice of parameters at initialization, while having no influence in the hydrodynamic limit since the output is 0 by construction.

\subsection{Explicit solution for an infinite training set}\label{ssec:linsepsol}
We first consider the limit of infinite number of examples, and later discuss the effects induced by a finite training set.  
\begin{figure}[!hbt]
   \centering
   \includegraphics[width=\columnwidth]{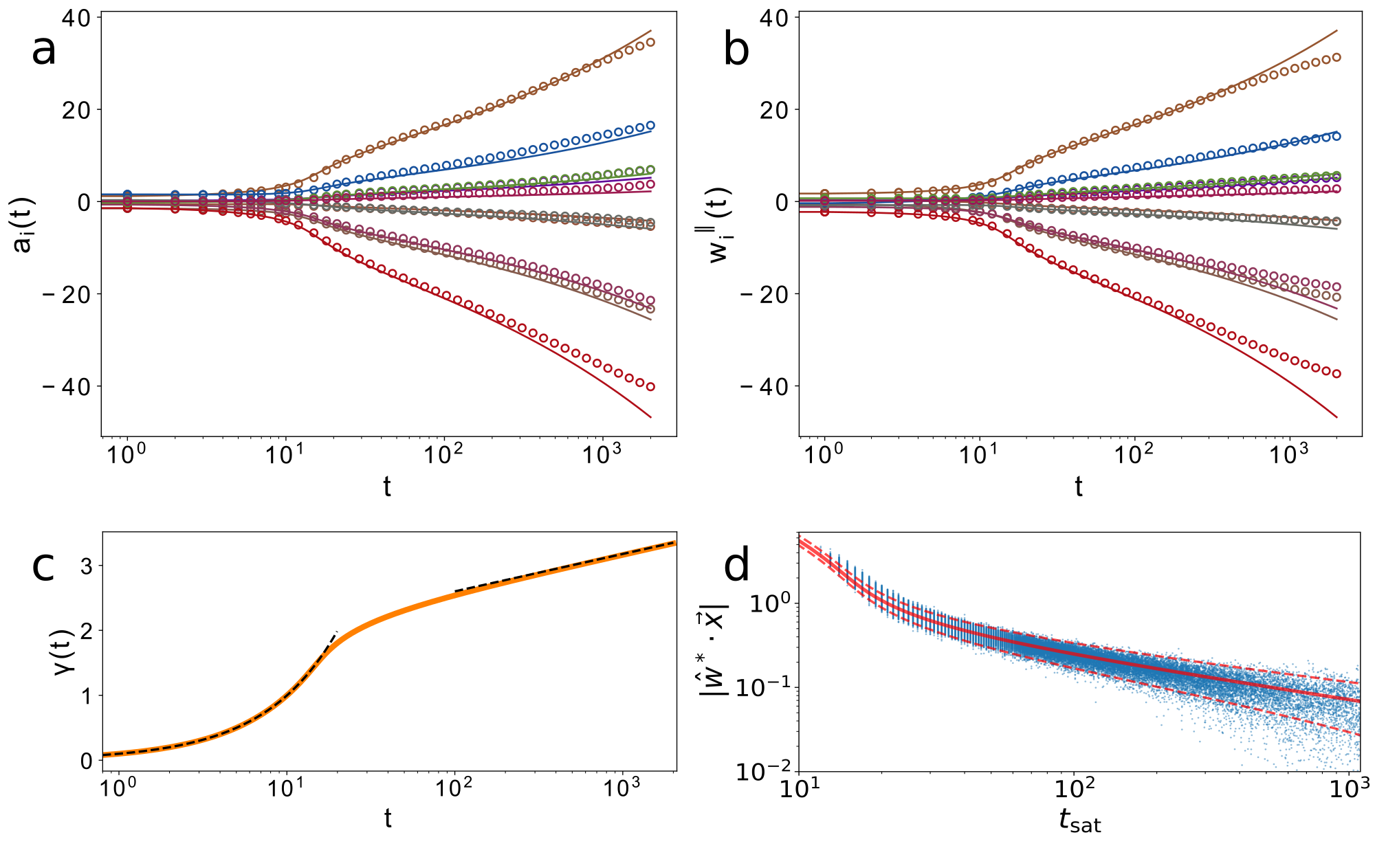}
   \caption{
    Training of a network with $M=400$, $N=10^5$, $d=100$, $\alpha=1.0$, $h=1$, $\beta^*=10^3$, for $t_{\rm max}=2\cdot10^3$ timesteps (until all examples are classified) with final generalization error $\sim0.01$ evaluated on $10^5$ examples. Data and initial parameters are taken from a normal distribution of zero mean and width 1 per dimension.
    {\bf a, b:} Evolution of ten of the $a_i(t)$s in (a) and of the $\wpar_i(t)$s in (b) during training (circles) compared to our theoretical prediction (lines) for the same initial values.
    {\bf c:} Evolution of $\gamma(t)$ obtained through numerical integration of eq.~\ref{eq:gamma} for the parameters of this example. The dashed lines represent the linear approximation near $t=0$ and the logarithmic slope $\log(t)/4$ for large $\gamma$ (shifted with a fitted constant).
    {\bf d:} Projection of examples on the vector $\wref$ as a function of the time $t_{\rm sat}$ when they are first satisfied. The red line is the estimate of our theory, the dashed lines represent our estimate for a standard deviation due to the finite number of nodes $M$ (see Sec.~\ref{ssec:finitesize}). 
   }\label{fig:training}
\end{figure}

The explicit solution of the training dynamics is obtained making use of the
cylindrical symmetry around $\wref$, which implies that 
\begin{equation}\label{eq:u}
    \avg{u(\xb,y;t)\theta\left({\wb\cdot\xb}\right)y\,\xb}_{\xb,y}=I(t)\wref.\label{csym}
\end{equation}
where $I(t)\equiv\avg{u(\xb,y;t)\theta\left({\wb\cdot\xb}\right)y\,\xb\cdot\wref }_{\xb,y}$. By plugging the identity (\ref{csym}) into eqs.~(\ref{gradl}, \ref{aweq}) one finds that the hydrodynamic equation (\ref{hydro}) can be solved
by the method of the characteristic, where  $\rho(\thb,t)$ is obtained by transporting the initial condition through the equations (\ref{aweq}). By decomposing the vector $\wb$ in its parallel and perpendicular components with respect to $\wref$, i.e.  $\wb=\wpar\wref+\mathbf{\wperr}$, and using the solution $\rho(\thb,t)$, one finds that
the parameters $\thb$ at time $t$ are distributed in law as:
\begin{equation}\label{eq:eqmot1}\left\{\begin{array}{rcl}
    a(t)&\stackrel{d}{\sim}&a(0)\cosh(\gamma(t))+\wpar(0)\sinh(\gamma(t))\\
    \wpar(t)&\stackrel{d}{\sim}&\wpar(0)\cosh(\gamma(t))+a(0)\sinh(\gamma(t))\\
    {\mathbf\wperr}(t)&\stackrel{d}{\sim}&{\mathbf\wperr}(0)\\
\end{array}\right.;\qquad
    \gamma(t)=\frac{\beta}{\alpha\sqrt{d}}\int_0^tI(t)\de t.
\end{equation}
where $a(0),\,\wpar(0),\,{\mathbf\wperr}(0)$ are given by the initial condition distributions, i.e. they are i.i.d Gaussian. 
Using the distribution of $\thb$ at time $t$, one can then compute 
$\avg{u(\xb,y;t)\theta\left({\wb\cdot\xb}\right)y\,\xb\cdot\wref }_{\xb,y}$ and hence obtain a self-consistent equation on $I(t)$, which 
completes the mean-field solution. Similarly, one can obtain explicitly the output function and the indicator function which acquire a simple form:
\begin{equation}\label{eq:fout}
    f(\xb;\thb)=\frac{\sinh(2\gamma(t))}{2\sqrt{d}}\wref \cdot\xb,
\end{equation}
\begin{equation}\label{uw}
    u(\xb,y;t)=\theta\left(\frac{2h\sqrt{d}}{\alpha\sinh(2\gamma(t))}-y\wref \cdot\xb\right)
    \end{equation}
where we have used that $f(\xb;\thb)=0$ at $t=0$.
As expected, both functions have cylindrical symmetry around $\wref$. The analytical derivation of these results and the following ones is presented in the Appendix~\ref{ssec:SMIt}.\\
Since by definition $I(t)\ge 0$ the function $\gamma(t)$ is monotonously increasing and starts from zero at $t=0$. To be more specific, we consider two cases: normally distributed data with unit variance in each dimension, and uniform data on the $d$-dimensional unit sphere. The corresponding self-consistent equations on $\gamma(t)$ read respectively:
  \begin{equation}\label{eq:gamma}
    \dot\gamma(t)=\frac{\beta I^N(0)}{\alpha\sqrt{d}} \left(1-\exp\left[-\frac{2h^2d}{\alpha^2\sinh^2(2\gamma(t))}\right]\right),
\end{equation}  
\begin{equation}\label{eq:gammasph}
    \dot\gamma(t)=\frac{\beta I^S(0)}{\alpha\sqrt{d}} \left(1-\max\left(0,1-4h^2d/(\alpha^2\sinh^2(2\gamma(t))) \right)^{\frac{d-1}{2}}\right),
\end{equation}
where $I^N(0)=1/\sqrt{2\pi}$  and $I^S(0)=\Gamma\left(\frac{d+2}{2}\right)/ (\Gamma\left(\frac{d+1}{2}\right)d\sqrt{\pi})$. Both equations imply that $\gamma(t)\sim t$ for small $t$ and $\gamma(t)\sim \ln t$ for large $t$. 

We have now gained a full analytical description of the training dynamics: the node parameters evolve in time following eqs.~(\ref{eq:eqmot1}). Note that their trajectory is independent of the training parameters and the initial distribution, which only affect the time dependence, i.e. the ``clock'' $\gamma(t)$.
The change of the output function is given by eq.~(\ref{eq:fout}), where one sees that only the amplitude of $f(x,\thb)$ varies with time and is governed by $\gamma(t)$. The amplitude increases monotonically so that more examples can be classified above the margin $h$ at later times; the more examples are classified the slower becomes the increase of $\gamma(t)$ and hence the dynamics. 

Our theoretical prediction can be directly compared with a simple numerical experiment.
Fig.~\ref{fig:training} shows the training of a network with $M=400$ on Gaussian input data. The top panels ({\bf a} and {\bf b}) compare the analytical evolution of the network parameters $a_i$ and $\wpar_i$ obtained from eqs.~(\ref{eq:eqmot1}) to the numerical one. In {\bf c} we plot $\gamma(t)$ (computed numerically) showing that it grows linearly in the beginning and logarithmically at longer times, as expected from theory. In {\bf d} we show a scatter plot illustrating that the time when an example is satisfied is proportional to its projection on the reference vector, following on average our estimate based on eq.~(\ref{uw}). 
Overall, the agreement with the analytical solution is very good. The spread around the analytical solution in panel {\bf d} is a finite-$M$ effect, that we will analyze in Sec.~\ref{ssec:finitesize}.
The departure from the analytical result (\ref{eq:eqmot1}) happens at large time when the 
finiteness of the training set starts to matter (the larger is the training set the larger is this time). 
In fact, for any finite number of examples the empirical average over unsatisfied examples deviates from its population average and the dynamics is modified eventually, and ultimately stops when the whole training set is classified beyond margin.
We study this regime in Sec.~\ref{ssec:finitesize}. 

%%%%%%%%%%%%%%%%%%%%%%%%
\subsection{Lazy learning and rich learning regimes}\label{ssec:lazy}

\begin{figure}[!hbt]
   \centering
   \includegraphics[width=\columnwidth]{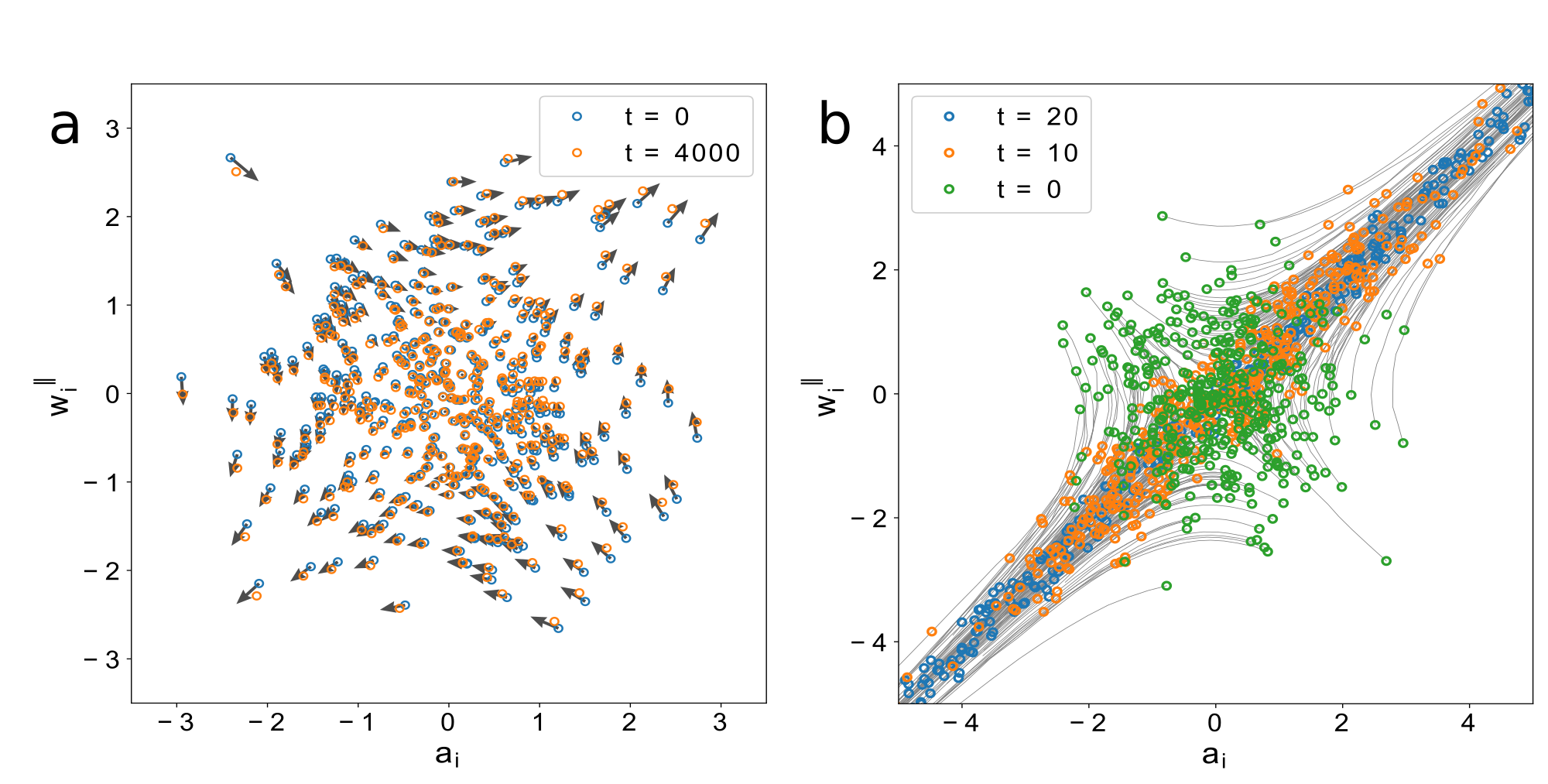}
   \caption{Evolution of $a_i$ and $\wpar_i$ for a network with $M=400$, $N=10^4$, $d=100$, $h=1$ in two different regimes. Data and initial parameters are taken from a normal distribution of zero mean and width 1 per dimension.
   {\bf a:} First and last step of a case with $\alpha=10^3$ (learning rate $\beta^*=10^4$, training set is fitted by $t=3000$, final generalization error $\sim0.04$). The arrows indicate the analytical derivative at $t=0$, showing that the evolution is approximately linear.
   {\bf b:} Initial steps (time indicated in legend) of a case with $\alpha=10^{-3}$ (learning rate $\beta^*=1$, training set is fitted by $t=300$, final generalization error $\sim0.02$). The gray lines follow the evolution of each node.
   }\label{fig:awevol}
\end{figure}

The presence of the factor $\alpha$ in the loss function~(\ref{eq:loss1}) allows us to explore explicitly the crossover between different learning regimes, in particular the {\it ``lazy learning''} regime corresponding to $\alpha\to\infty$~\cite{Chizat19}. The dynamical equations can be studied in this limit by introducing $\overline{\gamma}(t)=\alpha \gamma(t)$. For concreteness, let us focus on the case of normally distributed data. Taking the $\alpha \rightarrow \infty$ limit of eq.~(\ref{eq:gamma}) one finds the equation for $\overline{\gamma}(t)$:
  \begin{equation}\label{eq:gammabar}
    \dot{\overline{\gamma}}(t)=\frac{\beta I^N(0)}{\sqrt{d}} \left(1-\exp\left[-\frac{2h^2d}{4\overline{\gamma}(t)^2}\right]\right),
\end{equation}  
As for the evolution of the parameters and the output function, we obtain:
\begin{equation}\label{eq:eqmot1bis}\left\{\begin{array}{rcl}
    a_i(t)-a_i(0)&=&\ds\wpar_i(0)\frac{\overline{\gamma}(t)}{\alpha}+O(\alpha^{-2})\\[5pt]
    \wpar_i(t)-\wpar_i(0)&=&\ds a_i(0)\frac{\overline{\gamma}(t)}{\alpha}+O(\alpha^{-2})
\end{array}\right.;\qquad
    \alpha f(\xb;\thb)=\frac{\overline{\gamma}(t)}{\sqrt{d}}\wref \cdot\xb.
\end{equation}
The equations above provide an explicit solution of lazy learning dynamics and illustrate its main features: the $\thb_i$ evolves very little and along a fixed direction, in this case given by $(\wpar_i(0),a_i(0),0)$. Despite the small changes in the nodes parameters, of the order of $1/\alpha$, the network does learn since classification is performed through $\alpha f(\xb;\thb)$ which has an order one change even for $\alpha \rightarrow \infty$. 
In this regime, the correlation between $a$ and $\wpar$ only increases slightly, but this is enough for classification, since an infinite amount of displacements in the right direction is sufficient to solve the problem.\\
On the contrary, when $\alpha$ is of order one or smaller, the dynamics is in the so-called {\it ``rich learning''} regime~\cite{woodworth2020kernel}.
At the beginning of learning, the initial evolution of the ${\thb}_i$s follows the same linear trajectories of the lazy-learning regime. However, at later stages, the trajectories are no more linear and the norm of the 
weights increases exponentially in $\gamma(t)$, stopping only at very large values of $\gamma$ when all nodes are almost aligned with $\wref$ (for small $\alpha$).
Note that, as observed in~\citet{Geiger19}, with the standard normalization $1/\sqrt{M}$ it would be the parameter $\alpha\sqrt M$ governing the crossover between the two regimes.

We compare the two dynamical evolutions in Fig.~\ref{fig:awevol}. 
The left panel ({\bf a}) shows the displacement of parameters between initialization and full classification (zero training loss) for a network with $\alpha=10^3$. As expected, the displacement is small and linear. A very different evolution takes place for $\alpha=10^{-3}$ in the right ({\bf b}) panel. The trajectories are non-linear, and all nodes approach large values close to the $a=\wpar$ line at the end of the training. Correspondingly, the initially isotropic Gaussian distribution evolves towards one with covariance matrix $\cosh(2\gamma)$ on the diagonal and $\sinh(2\gamma)$ off diagonal.

Note that for all values of $\alpha$, even  very large ones, the trajectories 
of the $\thb_i$s are identical and given by eqs.~(\ref{eq:eqmot1}). What differs is the ``clock'' $\gamma(t)$, in particular for large $\alpha$ the system remains for a much longer time in the lazy regime. This is true as long as the number of training samples is infinite. Instead, if the number of data is finite, the dynamics stops once the whole training set is fitted: for large $\alpha$ this happens before the system is able to leave the lazy regime, whereas for small $\alpha$ a full non-linear (rich) evolution takes place. Hence, the finiteness of the training set leads to very distinct dynamics and profoundly different ``trained'' models (having both fitted the training dataset) with possibly different generalization properties~\cite{Lee19,Geiger19,Arora19b}.

%%%%%%%%%%%%%%%%%%%%%%%%
\subsection{Beyond mean-field theory}\label{ssec:finitesize}
The solution we presented in the previous sections holds in the limit of an infinite number of nodes and of training data. Here we study the corrections to this asymptotic limit, and discuss the new phenomena that they bring about.

\begin{figure}[!hbt]
   \centering
   \includegraphics[width=\columnwidth]{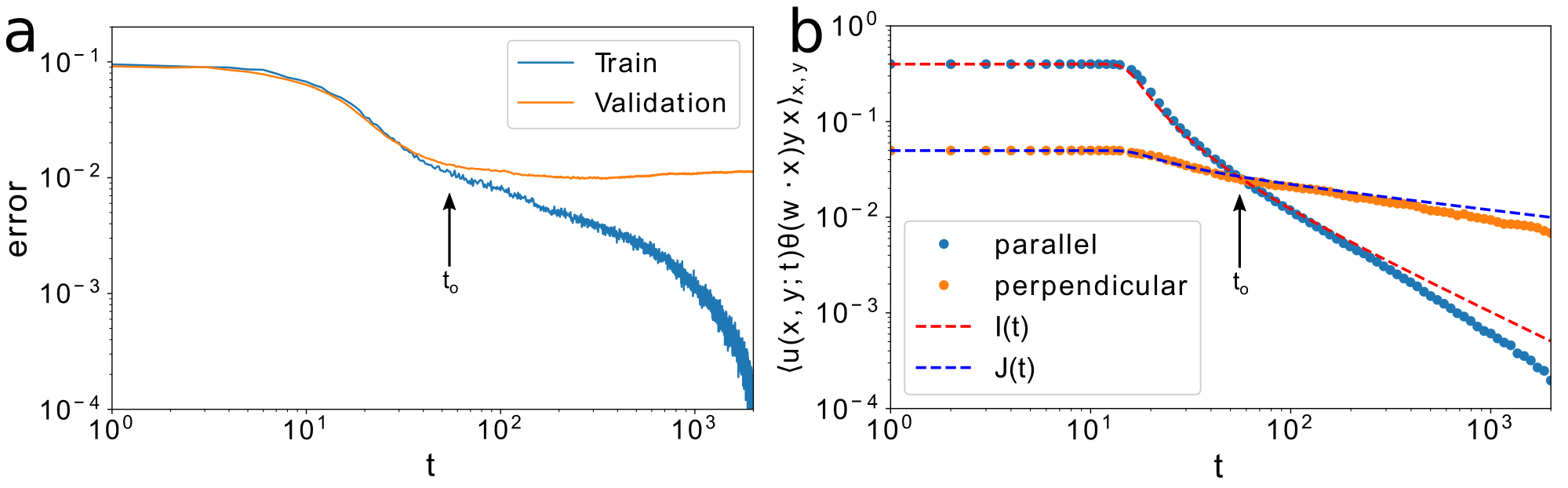}
   \caption{
   {\bf a:} Training (blue) and generalization (orange) error (fraction of misclassified examples), during training with same parameters as Fig.~\ref{fig:training}.
   {\bf b:} Components of $\avg{u(\xb,y;t)\theta\left({\wb\cdot\xb}\right)y\,\xb}_{\xb,y}$ along $\wref$ (parallel) and perpendicular to it, during training. The dots are numerical results for the same training show in {\bf a}. The lines represent our analytical predictions $I(t)$ and $J(t)/\sqrt N$ for the same parameters.
   }\label{fig:finite}
\end{figure}

{\bf Finite number of nodes.}
In the large $M$ limit the $a_i$ and ${\wb}_i$ are Gaussian i.i.d.\ variables. By the central limit theorem, the function (\ref{eq:f}) concentrates around its average, and has negligible fluctuations of the order of $1/\sqrt{M}$ when $M\rightarrow\infty$. If $M$ is large but finite (keeping an infinite training set), these fluctuations of $f(x,\thb)$ are responsible for the leading corrections to mean-field theory. In Appendix~\ref{ssec:SMfinite} we compute explicitly the variance of the output function, $\lim_{M\rightarrow \infty }M\textnormal{Var}[f(x,\thb)]=\sigma^2_f(t)$, with  
\begin{equation}\label{eq:sigf}
    \sigma_f^2(t)\equiv((5\cosh^2(2\gamma(t))-2\cosh(2\gamma(t))-3)(\wref\cdot\xb)^2+ 2\cosh(2\gamma(t))\abs{\xb}^2)/(4d)
\end{equation}
The main effect of this correction is to induce a spread in the dynamics, e.g.\ of the data with same satisfaction time. This phenomenon is shown in Fig.~\ref{fig:training}({\bf d}) for $M=400$, where we compare the numerical spread to an estimate of the values of $\wref\cdot\xb$ such that the hinge is equal to the average plus or minus one standard deviation (details on this estimate in Appendix~\ref{ssec:SMfinite}).

{\bf Finite number of data.}
We now consider a finite but large number of examples $N$ (keeping infinite the number of nodes).
In the large $N$ limit the empirical average over the data in $\avg{u(\xb,y;t)\theta\left({\wb\cdot\xb}\right)y\,\xb}_{\xb,y}$ converges to its mean $I(t)\wref$. The main effect of considering a finite $N$ is that the empirical average fluctuates around this value.  
Using the central limit theorem we show in Appendix~\ref{ssec:SMfinite} that the leading correction to the asymptotic result reads:
\begin{equation}\label{eq:Ncorr}
    \avg{u(\xb,y;t)\theta\left({\wb\cdot\xb}\right)y\,\xb}_{\xb,y}=I(t)\wref+\frac{J(t)}{\sqrt N}{\mathbf{\delta w}}_\perp+O(1/N) 
\end{equation}
where $\mathbf{\delta w}_\perp$ is a unitary random vector perpendicular to $\wref$ and $J(t)\equiv \sqrt{(d-1)f^U(t)/2}$.
The term $f^U(t)\equiv \avg{u(\xb,y;t)}_{\xb,y}$, the fraction of unsatisfied examples at time $t$, controls the strength of the correction, as expected since only unsatisfied data contribute to the empirical average $\langle \cdot \rangle_{\xb,y}$.
The vector on the RHS of (\ref{eq:Ncorr}) is the one towards which all the ${\wb}_i$ align, see eqs.~(\ref{eq:eqmot1}). Therefore, the main effect of the correction (\ref{eq:Ncorr}) is for the nodes parameters to align along a direction which is slightly different from $\wref$ and dependent on the training set. This naturally induces different accuracies between the training and the test sets, i.e.\ it leads to {\it overfitting}.\footnote{The two accuracies instead coincide for $N\rightarrow\infty$, since all possible data are seen during the training and no overfitting is present in the asymptotic limit.}
Note that the strength of the signal, $I(t)$, is roughly of the order of the fraction of unsatisfied data $f^U(t)$, whereas the noise due to the finite training set is proportional to the square root of it. The larger the time, the smaller $f^U(t)$ is, hence the stronger are the fluctuations with respect to the signal. 
In Fig.~\ref{fig:finite}({\bf b}) we compute numerically the components of 
$\avg{u(\xb,y;t)\theta\left({\wb\cdot\xb}\right)y\,\xb}_{\xb,y}$ parallel and perpendicular to $\wref$, and compare them to $I(t)$ and $J(t)/\sqrt{N}$. Remarkably, we find a very good agreement even for times when $J(t)/\sqrt{N}$ is no longer a small correction.
This suggest that an estimate of the time $t_{o}$ when overfitting takes place is given by $I(t_o)=J(t_o)/\sqrt N$. We test this conjecture in panel ({\bf a}): indeed the two contributions are of the same order of magnitude for $t_o\sim50$, which is around the time when training and validation errors diverge.

%%%%%%%%%%%%%%%%%%%%%%%%
\subsection{Mislabeling}\label{ssec:mislabel}

We now briefly address the effects due to noise in the labels, see Appendix~\ref{ssec:SMmisl} for detailed results and Appendix~\ref{ssec:SMdatamislab} for numerical experiments. Mislabeling is introduced by flipping the label of a small fraction $\delta$ of the examples. The main effect is to decrease the strength of the signal, $I(t)$, since the mislabeled data lead to an opposite contribution in (\ref{eq:u}) with respect to the correctly labeled ones. In the asymptotic limit of infinite $N$ and $M$, the reduction of the signal slows down the dynamics, which stops when the number of unsatisfied correct examples equals the one of mislabeled ones.
For large but finite $N$, the noise $J(t)/\sqrt N$ is enhanced with respect to the signal because its strength is related to the fraction of {\it all} unsatisfied examples, and not just the correctly labeled ones. 
Hence, overfitting is stronger and takes place earlier with respect to the case analyzed before.

%%%%%%%%%%%%%%%%%%%%%%%%%%%%%%%%%%%%%%%%%%%%%%%%%%%%%%%%%
\section{Discussion and Experiment}\label{sec:disc}
\begin{figure}[!hbt]
   \centering
  \includegraphics[width=\columnwidth]{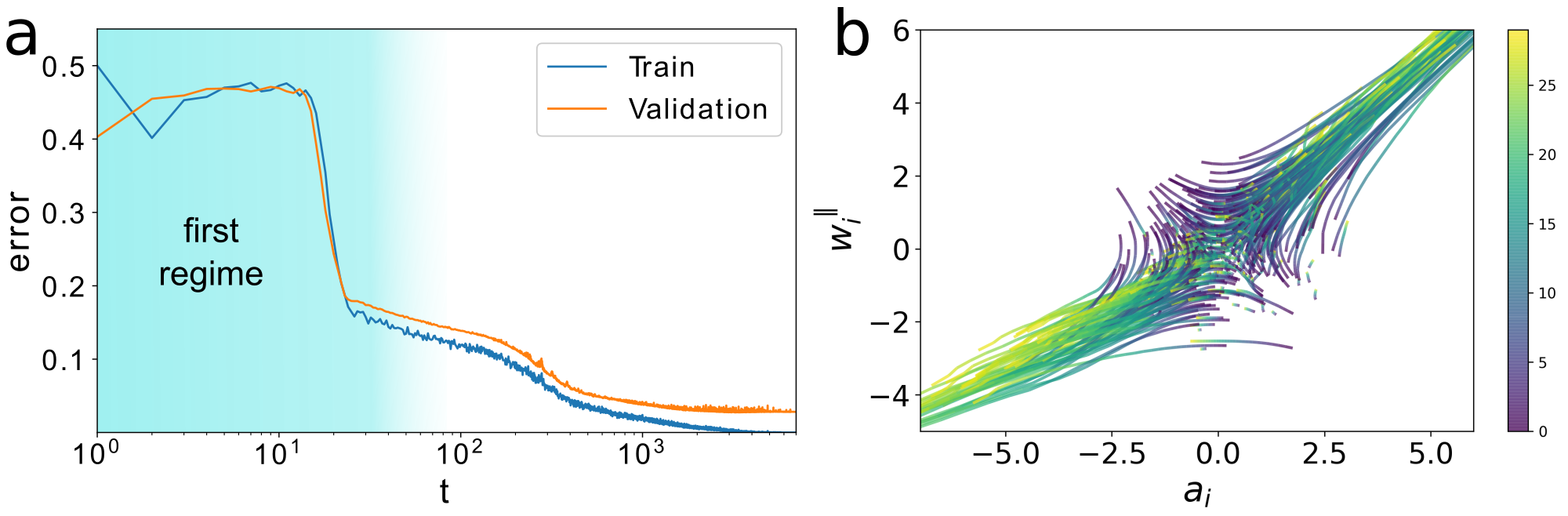}
   \caption{
   {\bf a:} Training (blue) and generalization (orange) error for a network with $M=400$, trained on $N=10^4$ MNIST data ($d=784$), with parity labels. 
   Inputs are only rescaled by a factor $1/255$, no further processing is done. The training is performed with $\beta^*=1000$, $\alpha=1$, $h=1$ and the validation error on $10^4$ examples is $\sim0.03$ after $2000$ evolution steps. The shaded area represents the area where our theory applies.
   {\bf b:} Evolution of $a_i$ and $\wpar_i$ in the first $30$ steps of training. The color (see color bar) represents the step of evolution. 
   %To compute $\wpar_i$, we take as $\wref$ the difference between the averages of the two parity sets (scaled to length 1).
   }\label{fig:MNIST}
\end{figure}

We have provided an analytical theory for the dynamics of a single hidden layer neural network trained for binary classification with linear hinge loss. We have found two dynamical regimes: a first one, correctly accounted for by mean-field theory, in which every node has its own dynamics with a time-dependent dataset determined self-consistently from the average nodes population. During this evolution the nodes parameters align with the direction of the reference classification vector. In the second regime, which is not accounted for by mean-field theory, the noise due to the finite training set becomes important and overfitting takes place. 
The merit of the model we focused on is that, thanks to its simplicity, several effects happening in real networks can be studied in detail analytically. Several works have shown distinct dynamical regimes in the training dynamics: first the network learns coarse grained properties, later on it captures the finer structure, and eventually it overfits~\cite{baity2019comparing,Goldt19,saad1996dynamics,bordelon2020spectrum}. Given the simplicity of the dataset we focused on, we expect our model to describe the first regime but not the second one, which would need a more complex model of data.
To test this conjecture, we train our network to classify the parity of MNIST handwritten digits~\cite{mnist}. To establish a relationship with our case, we define $\wref$ as the direction of the difference between the averages of the two parity sets. We can now define $\wpar$ for each node, and study the dynamics of $a_i,\wpar_i$.
We report in Fig.~\ref{fig:MNIST} the evolution of these parameters in the early steps of training, in which the training loss decreases of $65\%$ of its initial value (Fig.~\ref{fig:MNIST}{\bf a}). 
The evolution of the parameters (Fig.~\ref{fig:MNIST}{\bf b}) bears a strong resemblance with our findings, see the remarkable similarity with Fig.~\ref{fig:awevol}({\bf b}). 

%%%%%%%%%%%%%%%%%%%%%%%%%%%%%%%%%%%%%%%%%%%%%%%%%%%%%%%%%
% \section*{Broader Impact}
% Given the purely theoretical scope of this paper, it does not seem to present any foreseeable societal consequence.

%%%%%%%%%%%%%%%%%%%%%%%%%%%%%%%%%%%%%%%%%%%%%%%%%%%%%%%%%
\begin{ack}
We thank S. d'Ascoli and L. Sagun for discussions, and M. Wyart for exchanges about his work on a similar model \cite{preprintW}. \\\\
We acknowledge funding from the French government under management of Agence Nationale de la Recherche as part of the ``Investissements d’avenir'' program, reference ANR-19-P3IA-0001 (PRAIRIE 3IA Institute) and from the Simons Foundation collaboration ``Cracking the Glass Problem'' (No. 454935 to G. Biroli). 
\end{ack}

%%%%%%%%%%%%%%%%%%%%%%%%%%%%%%%%%%%%%%%%%%%%%%%%%%%%%%%%%
\appendix

\section{Explicit calculations}

\subsection{Derivation of the hydrodynamics mean-field equation}\label{ssec:SMhydro}
In order to simplify the derivation in the following we use a compact notation 
for the function $f$:
\begin{equation}
    f({\xb;\thb})=\frac 1 M \sum_{i=1}^M \overline{\sigma}({\xb;\thb_i})
\end{equation}
where $\overline{\sigma}({\xb;\thb_i})\equiv a_i \sigma\left(\frac{\wb_i\cdot\xb}{\sqrt{d}}\right)$, and for the gradient flow equations on the parameters of the network:
\begin{equation}
   \dot \thb_i=-\frac{\beta}{N}\sum_{n=1}^N
    \frac{\partial \ell(y_n,f({\xb_n;\thb}))}{\partial f}
    \frac{\partial \overline{\sigma}({\xb_n;\thb_i})}{\partial \thb_i}.
\end{equation}
The strategy to derive hydrodynamics mean-field equations developed in physics consists in using the following equation, valid for $M\rightarrow \infty$ and any test function $H$: 
\begin{equation}
    \frac 1 M \sum_{i=1}^M H(\thb_i(t))=\int d\thb H(\thb)\rho(\thb,t) 
\end{equation}
and then in differentiating RHS and LHS with respect to time, see e.g.~\cite{Dean96}. The important point here (and later) is that the density $\rho(\thb,t)$, which depends on the random initial conditions, concentrates in the large $M$ limit due to the  nature of the interaction between parameters, which is only through the function $f$, and the type of distributions considered for the initial conditions.\footnote{These two features lead to mean-field interactions in which one parameter interacts weakly with all the others.  In physical systems a particle instead  interacts only with a finite number of other particles, hence the density field remains highly fluctuating. Only performing coarse-graining in space and time one can get hydrodynamic equations, see~\cite{spohn2012large} for a rigorous presentation and \cite{chaikin1995principles} for a more general one.} 
The derivative of the RHS leads to 
\begin{equation}\label{rhs}
   \int d\thb H(\thb) \partial_t\rho(\thb,t) 
\end{equation}
whereas the derivative of the LHS reads:
\begin{equation}
   -\frac \beta M \sum_{i=1}^M \nabla_{\thb}H(\thb_i(t))
   \frac{1}{N}\sum_{n=1}^N\frac{\partial \ell(y_n,f({\xb_n;\thb}))}{\partial f} \nabla_{\thb}\overline{\sigma}({\xb_n;\thb_i(t)}). 
\end{equation}
We now use the identity:
\begin{equation}
  \frac{\delta {\mathcal L[\rho(\thb,t)]}}{\delta \rho(\thb,t)}= \frac{1}{N}\sum_{n=1}^N\frac{\partial \ell(y_n,f({\xb_n;\thb}))}{\partial f} \overline{\sigma}({\xb_n;\thb(t)}) 
\end{equation}
to rewrite the LHS as 
\begin{equation}
   -\frac \beta M \sum_{i=1}^M \nabla_{\thb}H(\thb_i(t))\nabla_{\thb_i(t)}\left.\frac{\delta {\mathcal  L[\rho(\thb,t)]}}{\delta \rho(\thb,t)}\right|_{\thb=\thb_i(t)}.
\end{equation} 
For $M\rightarrow \infty$ this expression can be rewritten as 
\begin{equation}\label{lhs}
    -\beta \int d\thb \rho(\thb,t) 
    \nabla_{\thb}H(\thb)\nabla_{\thb}\frac{\delta {\mathcal  L[\rho(\thb,t)]}}{\delta \rho(\thb,t)}= \int d\thb 
   H(\thb)\left[\beta\nabla_{\thb} \left(\rho(\thb,t)\nabla_{\thb}\frac{\delta {\mathcal L[\rho(\thb,t)]}}{\delta \rho(\thb,t)}\right)\right]
\end{equation}
where we have used an integration by part to obtain the last identity. Since the expressions in (\ref{rhs}) and (\ref{lhs}) are equal for any test function $H$, we obtain that the density $\rho(\thb,t)$  verifies Eq.~\ref{hydro} from the main text:
\begin{equation}
\partial_t\rho(\thb,t)=\beta\nabla_{\thb} \left(\rho(\thb,t)\nabla_{\thb}\frac{\delta {\mathcal L[\rho(\thb,t)]}}{\delta \rho(\thb,t)}\right)\,\,,\,\, \rho(\thb,0)=\gauss(0,\mathbb{I}).
\end{equation}
The initial condition for $\rho(\thb,t)$ is a Gaussian distribution since the parameters at initialization are i.i.d.\ Gaussian variables. 

\subsection{Calculation of $I(t)$}\label{ssec:SMIt}
We want to compute the integral of Eq.~(\ref{eq:u}) of the main text:
\begin{equation}
    \avg{u(\xb,y;t)\theta\left({\wb\cdot\xb}\right)y\,\xb}_{\xb,y}=
    \int \sum_{y=\pm1}u(\xb,y;t)\theta\left({\wb\cdot\xb}\right)y\,\xb P(\xb, y)\de \xb
\end{equation}
for the task and distributions mentioned in the text.

Let us start by observing that since $P(\xb,y)$ has spherical symmetry and $u(\xb,y;t)$ has cylindrical symmetry around $\wref$ and is symmetric under inversion along $\wref$ (because of the label symmetry of the problem), the whole integrand without the $\theta\left({\wb\cdot\xb}\right)$ is symmetric under inversion operation.
Indeed, $P(\xb,y)=P(-\xb,-y)$, $u(\xb,y;t)=u(-\xb,-y;t)$ and $y\xb={\rm sign} (\wref\cdot\xb)\xb={\rm sign} (\wref\cdot(-\xb))(-\xb)$.
The effect of the $\theta\left({\wb\cdot\xb}\right)$ term is to select one particular half-space over which the integral is done. However, because of the symmetric under inversion 
the integral on {\it any} half space is equivalent, hence the result is independent of $\wb$.
Moreover for any direction 
%$\wb_\perp$ 
orthogonal to $\wref$, the integrand is odd under inversion of that component, and is therefore $0$. The only component different from zero is then the one along $\wref$, dubbed $I(t)$ in the text. Let us define $\wref\cdot\xb=\xpar$ and notice that that $y\xpar={\rm sign} (\xpar)\xpar =\abs{\xpar}$ so that we can for simplicity consider the integral on the positive values
\begin{equation}\label{eq:It1}
    I(t)=\int_{\xpar>0} \sum_{y=\pm1}u(\xb,y;t)\xpar P(\xb, y)\de \xb.
\end{equation}

We will now consider the specific expression found in the main text $u(\xb,y;t)=\theta(H-y\xpar)$, and for the noiseless case $P(\xb, y)=P(\xb)\theta(y\xpar)$.

In the case of normally distributed data, all orthogonal directions integrate to $1$ and we are left with a simple Gaussian integral
\begin{equation}
    I(t)=\int_0^H \xpar \gauss_{0,1}(\xpar)\de \xpar=\frac{1}{\sqrt{2\pi}}
    \left(1-e^{-H^2/2}\right).
\end{equation}
With $H=\frac{2h\sqrt{d}}{\alpha\sinh(2\gamma(t))}$ and $\dot\gamma(t)=\frac{\beta}{\alpha\sqrt{d}}I(t)$, we recover Eq.~(\ref{eq:gamma}) from the text.

For the case of data uniformly distributed on the $d-1$-dimensional unit sphere in $d$ dimensions, we divide by the sphere surface $S_{d-1}$ and integrate on the $d-1$ angular coordinates. Because of the symmetry, we perform $d-2$ angular integrals and obtain the surface of the $d-2$-dimensional sphere. The $u(\xb,y;t)=\theta(H-y\xpar)$ limit will set the extreme of integration to $\arccos(H)$ for $H<1$ and not affect the integral otherwise. Considering for simplicity directly the $H<1$ limit we obtain:
\begin{equation}
    I(t)=\frac{S_{d-2}}{S_{d-1}}\int_{\arccos H}^{\pi/2} \cos(\phi)\sin^{d-2}(\phi)\de\phi=\frac{S_{d-2}}{(d-1)S_{d-1}}
    \left[1-\left(\sin\arccos(H)\right)^{d-1}\right].
\end{equation}
Using the equation $S_{n-1}=n\pi^{n/2}/\Gamma(n/2+1)$ for the sphere surface and properly accounting for the different $H$ cases we recover Eq.~(\ref{eq:gammasph}) from the text.

\subsection{Calculation of finite size quantities}\label{ssec:SMfinite}
{\bf Finite number of nodes.}
To estimate the fluctuations due to a finite number of nodes, we will have to estimate the width of the output distribution for a given set of parameters.
Let us explicit from Eqs.~(\ref{eq:eqmot1}) of the main text for the parameters evolution that, starting from i.i.d.\ Gaussian initialization, the distribution of $(a,\wpar)$ is
\begin{equation}
    \rho(a(t),\wpar(t))=\gauss\left(0,\left(
    \begin{array}{cc}
         \cosh(2\gamma(t))&\sinh(2\gamma(t))  \\
         \sinh(2\gamma(t))&\cosh(2\gamma(t)) 
    \end{array}\right)\right),
\end{equation}
while all perpendicular components remain i.i.d.

The average output $f(\xb;\thb)$ for an example $\xb$ can then be simply computed from its definition as
\begin{equation}
    \int_{-\infty}^\infty\de a(t)\int_0^\infty \de\wpar(t)\frac{a(t)\wpar(t)\xpar}{\sqrt{d}}\rho(a(t),\wpar(t))=
    \frac{\xpar}{2\sqrt{d}}\avg{a(t)\wpar(t)}=
    \frac{\sinh(2\gamma(t))\xpar}{2\sqrt{d}}
\end{equation}
(all orthogonal integrals being equal to 1), having defined again $\wref\cdot\xb=\xpar$.
This proves Eq.~(\ref{eq:fout}) of the main text.

In order to estimate the fluctuations we should however compute the integral (we drop the t dependence for simplicity)
\begin{equation}
    \avg{f(\xb;\thb)^2}_\thb=
    \frac{1}{d}\int\de a\,\de\wb\,a^2(\wb\cdot\xb)^2\,\theta(\wb\cdot\xb)\rho(a,\wb).
\end{equation}
Since the integral is 1 for any direction perpendicular to $\xb$, this is more easily done considering the distribution of $\wx=\wb\cdot\xhat$ (with $\xhat=\xb/\abs{\xb}$). 
Defining $\wplh$ as $(\xb-\xpar\wref)/\abs{\xb-\xpar\wref}$, i.e.\ the versor in the direction of $\xhat$ perpendicular to $\wref$, we can write
$\xhat=\cos\theta\wref+\sin\theta\wplh$ and calling $\wpl=\wb\cdot\wplh$ (being a component perpendicular to $\wref$ and therefore i.i.d) we can explicit 
$\wx=\wpar\cos\theta+\wpl\sin\theta$.

We can thus write the distribution for this component as
\begin{equation}
    \rho(a(t),\wx(t))=\gauss\left(0,\left(
    \begin{array}{cc}
         \cosh(2\gamma(t))&\sinh(2\gamma(t))\cos\theta  \\
         \sinh(2\gamma(t))\cos\theta&\cosh(2\gamma(t))\cos^2\theta+\sin^2\theta 
    \end{array}\right)\right),
\end{equation}
and the integral as just 
\begin{equation}\begin{array}{l}
    \ds\frac{\abs{\xb}^2}{d}\int\de a\,\de\wx\,a^2\wx^2\,\theta(\wx)\rho(a,\wx)=\\
    =\ds\frac{\abs{\xb}^2}{2d}\left(\cosh^2(2\gamma(t))\cos^2\theta+
    \cosh(2\gamma(t))\sin^2\theta+2\sinh^2(2\gamma(t))\cos^2\theta\right).
\end{array}\end{equation}

The total spread due to this is thus
\begin{equation}\begin{array}{l}
    \ds\sigma_f^2(t)\equiv\avg{f(\xb;\thb)^2}_\thb-\avg{f(\xb;\thb)}^2_\thb=\\
    \ds=\frac{\abs{\xb}^2}{4d}\left[\left((5\cosh^2(2\gamma(t))-2\cosh(2\gamma(t))-3\right)\cos^2\theta+2\cosh(2\gamma(t))\right],
\end{array}\end{equation}
which is equivalent to Eq.~(\ref{eq:sigf}) in the main text.

To estimate the error in Fig.~\ref{fig:training}({\bf d}) of the main text, we ask what are the values of $\xpar=\xb\cos \theta$ such that the average output plus or minus a standard deviation, divided by $\sqrt{M}$, would be equal to the threshold.
Since the standard deviation involves $\abs{\xb}^2$, we estimate its average value for points with a given $\xpar$, i.e.\ $\avg{\abs{\xb}^2|_{\xpar}}=\xpar^2+d-1$. The variance is thus the sum of two terms: $\sigma_\parallel^2=\left((5\cosh^2(2\gamma(t))-3\right)/(4dM)$ multiplying $\xpar^2$ and a constant $\sigma_0^2=(d-1)\cosh(2\gamma(t))/(2dM)$.
%{\color{red} Why there is no $-2 cosh $ in $\sigma_\parallel$?}
%\blue{Because it gets canceled by the part of the last term multiplied by $\xpar^2$}{\color{red} OK, got it}
Requesting that $h/\alpha=\sinh(2\gamma(t))\xpar_\pm/(2\sqrt{d})\pm\sqrt{\sigma_\parallel^2\xpar_\pm^2+\sigma_0^2}$ we find:
\begin{equation}
    \xpar_\pm=\frac{1}{\sinh^2(2\gamma(t))/(4d)-\sigma_\parallel^2}\left[
    \frac{h\sinh(2\gamma(t))}{2\alpha\sqrt{d}}\pm\sqrt{\frac{h\sigma_\parallel^2}{\alpha^2}
    +\frac{\sigma_0^2\sinh^2(2\gamma(t))}{4d}-\sigma_0^2\sigma_\parallel^2}\right].
\end{equation}
These values are the dashed lines reported in Fig.~\ref{fig:training}({\bf d}).

\begin{figure}[!hbt]
  \centering
  \includegraphics[width=0.7\columnwidth]{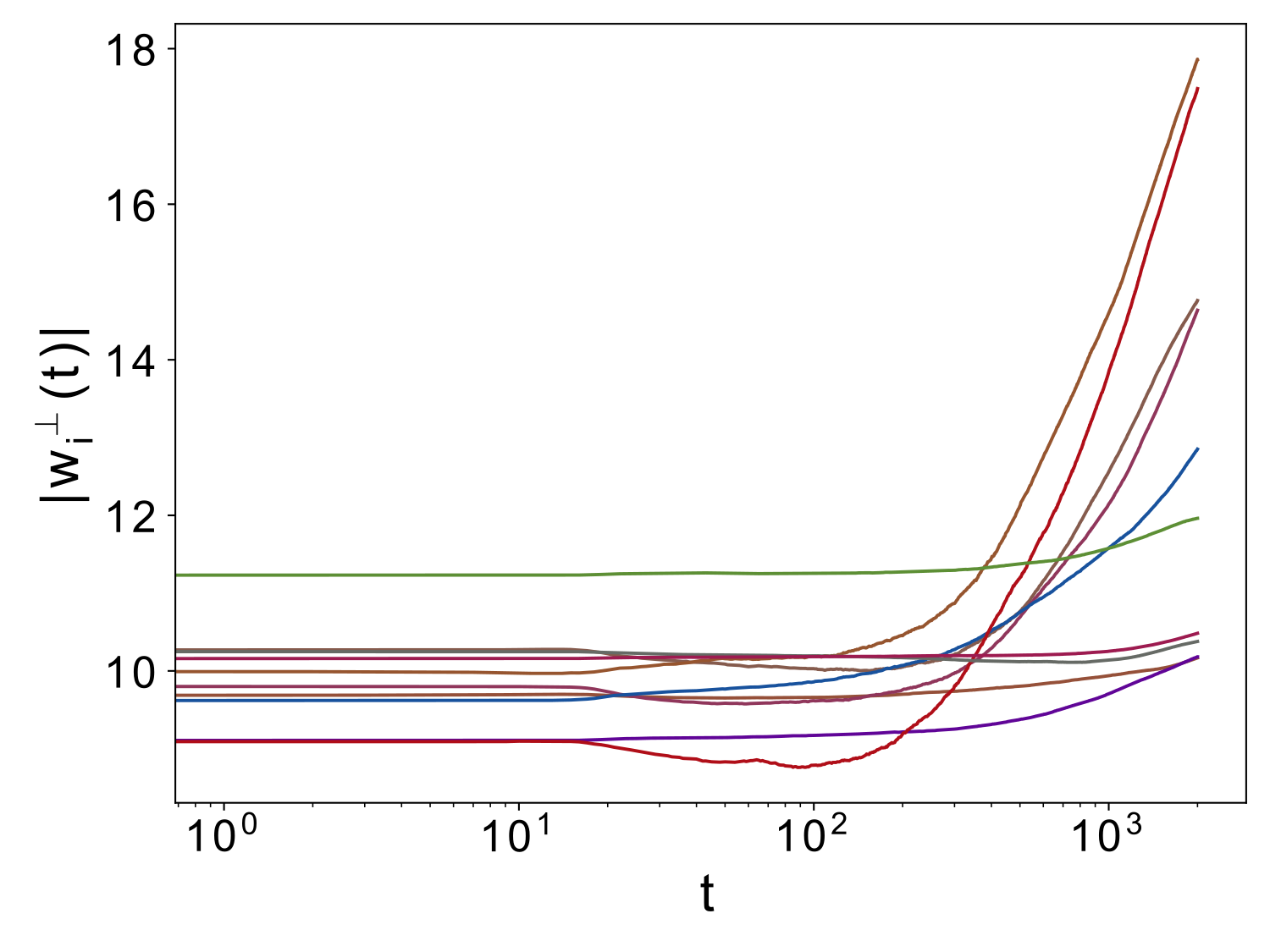}
  \caption{Evolution of $\abs{\wb_\perp(t)}$ for the same evolution as Fig.~1 of the main text.
  }\label{fig:wperp}
\end{figure}

{\bf Finite number of data.}
To estimate the fluctuations due to finite number of data in $\avg{u(\xb,y;t)\theta\left({\wb\cdot\xb}\right)y\,\xb}_{\xb,y}$ in the direction perpendicular to $\wref$, we use the central limit theorem, which gives fluctuations of the order  $\avg{\left(u(\xb,y;t)\theta\left({\wb\cdot\xb}\right)y\,\xb\right)^2}_{\xb,y}/{N}$.
We refer to section $\ref{ssec:SMIt}$ for the general symmetry considerations about that integral: in the case of normally distributed data, and if all data are not satisfied, i.e. $u(\xb,y;t)=1$ inside the empirical average over data, then for any given direction orthogonal to $\wref$ one obtains $1/{2N}$. Since there are $d-1$ such direction, this means that 
that considering finite number of data leads to a fluctuating component orthogonal to $\wref$ of norm of the order of $\sqrt{(d-1)/(2N)}$.

Let us consider now the case in which 
only $N^U=f^UN$ examples remain to satisfy, then the number of terms in the empirical sum is $N^U$ instead of $N$. In consequence, we obtain the same results than previously for the variance, but with an extra-factor $f^U$ in front, thus leading to an error of order $\sqrt{(d-1)f^U/(2N)}\equiv J(t)/\sqrt{N}$.

Estimating $f^U(t)$ for normally distributed data, and with the specific expression $u(\xb,y;t)=\theta(H-y\xpar)$ is then a simple Gaussian integral:
\begin{equation}
    f^U(t)\equiv\avg{u(\xb,y;t)}_{\xb,y}=
    \int_{-H}^H \gauss_{0,1}(\xpar)\de \xpar=\erf\left({\frac{H}{\sqrt{2}}}\right).
\end{equation}

Computing this for normally distributed data leads to:
\begin{equation}
    J(t)=\sqrt{\frac{(d-1)}{2}\erf\left(\frac{h\sqrt{2d}}{\alpha\sinh(2\gamma(t))}\right)},
\end{equation}
as was used to compute the estimates in  Fig.~\ref{fig:finite}({\bf b}) in the main text.

\subsection{Calculations for the mislabeling case}\label{ssec:SMmisl}
We now analyze the case, qualitatively described in the text, where a small fraction $\delta$ of the examples has been mislabeled as belonging to the opposite class.

Looking back at Eq.~(\ref{eq:It1}) and with $u(\xb,y;t)=\theta(H-y\xpar)$, it is clear that with an infinite number of examples the mislabeled ones are simply never classified, so that the $1-\delta$ fraction of correct examples gives rise to a normal dynamics, while the $\delta$ fraction of opposite ones contributes an opposite term of constant magnitude.
The effective integrals entering the dynamics are thus in this case
\begin{equation}\label{eq:misIt}
    I_\delta(t)=(1-\delta)I(t)-\delta I(0),
\end{equation}
and would drive the dynamics until the two contributions are equal.

When considering a finite number of data, as discussed in Sec.~\ref{ssec:SMfinite}, the number of unsatisfied examples with the correct label amounts to $(1-\delta)f^U(t)$, but since all the mislabeled examples are unsatisfied the total number will be incremented by $\delta$ leading to $f^U_\delta(t)=(1-\delta)f^U(t)+\delta$.

Again, evaluating this for the normally distributed case we find:
\begin{equation}\label{eq:misJt}
    J_\delta(t)=\sqrt{\frac{(d-1)}{2}\left[(1-\delta)\erf\left(\frac{\sqrt{2d}}{\alpha\sinh(2\gamma(t))}\right)+\delta\right]}.
\end{equation}

\begin{figure}[!bt]
  \centering
  \includegraphics[width=\columnwidth]{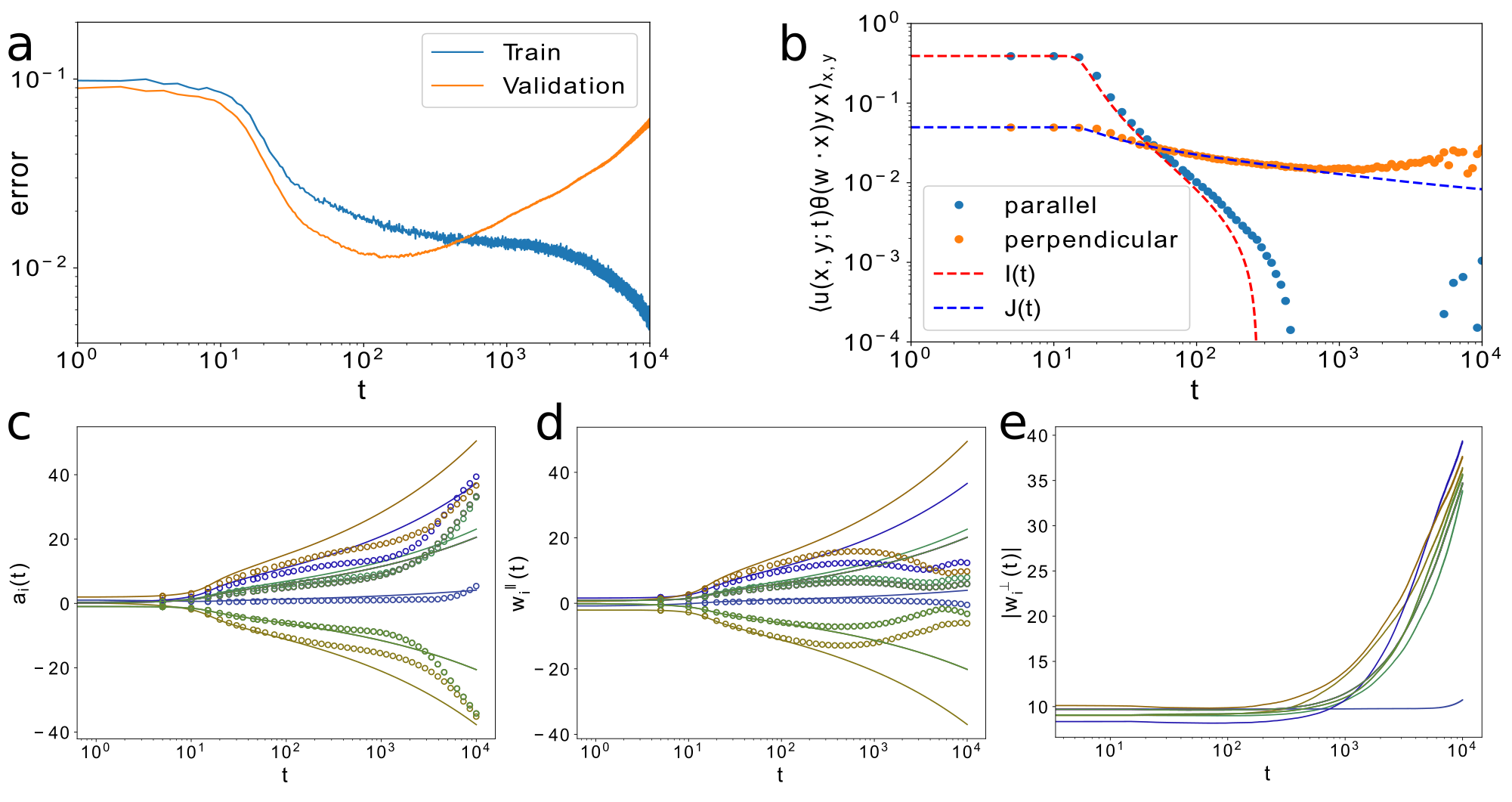}
  \caption{{
  \bf a:} Training (blue) and generalization (orange) error (fraction of misclassified examples), during a training with a small fraction $\delta=0.01$ of mislabeled examples. Training parameters: $M=400$, $N=10^5$, $d=100$, $\alpha=1.0$, $\beta^*=10^3$, timesteps $t_{\rm max}=10^4$, validation on $10^5$ examples.
  {\bf b:} Components of $\avg{u(\xb,y;t)\theta\left({\wb\cdot\xb}\right)y\,\xb}_{\xb,y}$ along $\wref$ (parallel) and perpendicular to it, during training. The dots are numerical results for the same training show in {\bf a}. The lines represent our analytical predictions $I_\delta(t)$ and $J_\delta(t)$ for the same parameters (Eqs. (\ref{eq:misIt}) and (\ref{eq:misJt})).
  {\bf c, d:} Evolution of a sample (10) of the $a_i(t)$ (c) and $\wpar_i(t)$ (d) during training (circles) compared to our theoretical prediction (lines) for the noiseless case with the same initial values and parameters.
  {\bf e:} Evolution of $\abs{\wb_\perp(t)}$ for the same sample of nodes.
  }\label{fig:mislab}
\end{figure}

%%%%%%%%%%%%%%%%%%%%%%%%%%%%%%%%%%%%%%%%%%%%%%%%%%%%%%%%%
\section{Further numerical experiments}

\subsection{Evolution of $\wb_\perp$}\label{ssec:SMwperp}
We report in Fig.~\ref{fig:wperp} the perpendicular component of the weights for a selection of nodes for the same example shown in Fig.~\ref{fig:training} of the main text.
As expected, the perpendicular component does not evolve for most of the training, and only increases moderately when we move into the overfitting regime.

\subsection{Quantities for the mislabeling case}\label{ssec:SMdatamislab}
We report here in Fig.~\ref{fig:mislab} some of the same quantities shown in Fig.~\ref{fig:training} and Fig.~\ref{fig:finite} of the main text, for a case where a small fraction $\delta=0.01$ of the examples are mislabeled.
As discussed in the main text, we can see how the dynamics still follows our estimate initially, then diverges into a much stronger overfitting state.
Panel {\bf b} shows a comparison of numerical quantities to our estimates of Sec.~\ref{ssec:SMmisl}: our estimate are still accurate up to the overfitting regime, after which the dynamics changes qualitatively.

%%%%%%%%%%%%%%%%%%%%%%%%%%%%%%%%%%%%%%%%%%%%%%%%%%%%%%%%%
\section{Other material}
\paragraph{Code.}
The code to reproduce all numerical results and graphs reported in this article can be found at \url{https://github.com/phiandark/DynHingeLoss/}. 
It consists of a single Jupyter notebook, based on Python 3 and requiring libraries numpy, scipy, tensorflow (1.xx), and matplotlib.
All examples can be run in a few minutes on a moderately powerful machine.
For more details, please see comments in the code.

\paragraph{Time evolution.}
An animation showing the training and validation error and parameters evolution for the same cases reported in Fig.~\ref{fig:awevol} can be found on the same page.
As discussed in Sec.~\ref{ssec:lazy}, the different behavior of the parameters is apparent, despite the similar final error. Moreover, the effects of overfitting can be noticed in the final phases of training.

%%%%%%%%%%%%%%%%%%%%%%%%%%%%%%%%%%%%%%%%%%%%%%%%%%%%%%%%%
\bibliographystyle{unsrtnat}
\bibliography{DynHingeLossArXiv}
\end{document}